\documentclass[sigconf]{acmart}

\usepackage{graphicx}
\usepackage{epstopdf}
\usepackage{booktabs} 
\usepackage{amsmath}
\usepackage{bm}
\usepackage{algorithmic}
\usepackage{algorithm}
\usepackage{amsmath}
\usepackage{multirow}
\usepackage{indentfirst}
\usepackage{subfigure}
\usepackage{epsfig}
\usepackage{epstopdf}
\usepackage{xspace}

\usepackage{makecell,rotating}

\usepackage{dsfont}

\usepackage{subeqnarray}
\newtheorem{myDef}{Definition}

\newcommand{\paratitle}[1]{\vspace{1.5ex}\noindent\textbf{#1}}
\newcommand{\ie}{\emph{i.e.,}\xspace}

\newcommand{\etal}{\emph{et al.}\xspace}

\newcommand{\name}{{X-Ensemble}\xspace}
\newcommand{\dname}{{X-Det}\xspace}

\newcommand{\ignore}[1]{}

\ignore{
\algnewcommand\algorithmicprocedure{\textbf{procedure}}
\algnewcommand\PROCEDURE{\item[\algorithmicprocedure]}%
\algnewcommand\algorithmicendprocedure{\textbf{end procedure}}
\algnewcommand\ENDPROCEDURE{\item[\algorithmicendprocedure]}%
\algnewcommand{\algvar}[1]{{\text{\ttfamily\detokenize{#1}}}}
\algnewcommand{\algarg}[1]{{\text{\ttfamily\itshape\detokenize{#1}}}}
\algnewcommand{\algproc}[1]{{\text{\ttfamily\detokenize{#1}}}}
\algnewcommand{\algassign}{\leftarrow}
}

\AtBeginDocument{%
  \providecommand\BibTeX{{%
    \normalfont B\kern-0.5em{\scshape i\kern-0.25em b}\kern-0.8em\TeX}}}

\copyrightyear{2020}
\acmYear{2020}
\setcopyright{acmcopyright}\acmConference[KDD '20]{Proceedings of the 26th ACM SIGKDD Conference on Knowledge Discovery and Data Mining}{August 23--27, 2020}{Virtual Event, CA, USA}
\acmBooktitle{Proceedings of the 26th ACM SIGKDD Conference on Knowledge Discovery and Data Mining (KDD '20), August 23--27, 2020, Virtual Event, CA, USA}
\acmPrice{15.00}
\acmDOI{10.1145/3394486.3403044}
\acmISBN{978-1-4503-7998-4/20/08}

\begin{document}
\fancyhead{}
\title{Interpretability is a Kind of Safety: \\An Interpreter-based Ensemble for Adversary Defense}

\author{Jingyuan Wang$^{1,3,4}$, Yufan Wu$^1$, Mingxuan Li$^1$, Xin Lin$^1$, Junjie Wu$^{2,3,\ast}$, Chao Li$^{1,4}$}
\affiliation{%
  \institution{$1.$ MOE Engineering Research Center of ACAT, School of Computer Science Engineering, Beihang University}
  \institution{$2.$ Beijing Key Laboratory of ESSTCO, School of Economics and Management, Beihang University, Beijing, China}
  \institution{$3.$ State Key Laboratory of Software Development Environment, Beihang University, Beijing 100191, China}
  \institution{$4.$ Beijing Advanced Innovation Center for BDBC, Beihang University, Beijing, China $\ast$ Corresponding author.}}

\renewcommand{\shortauthors}{J. Wang, et al.}

\begin{abstract}
While having achieved great success in rich real-life applications, deep neural network (DNN) models have long been criticized for their vulnerability to adversarial attacks. Tremendous research efforts have been dedicated to mitigating the threats of adversarial attacks, but the essential trait of adversarial examples is not yet clear, and most existing methods are yet vulnerable to hybrid attacks and suffer from counterattacks. In light of this, in this paper, we first reveal a gradient-based correlation between sensitivity analysis-based DNN interpreters and the generation process of adversarial examples, which indicates the Achilles's heel of adversarial attacks and sheds light on linking together the two long-standing challenges of DNN: fragility and unexplainability. We then propose an interpreter-based ensemble framework called X-Ensemble for robust adversary defense. X-Ensemble adopts a novel detection-rectification process and features in building multiple sub-detectors and a rectifier upon various types of interpretation information toward target classifiers. Moreover, X-Ensemble employs the Random Forests (RF) model to combine sub-detectors into an ensemble detector for adversarial hybrid attacks defense. The non-differentiable property of RF further makes it a precious choice against the counterattack of adversaries. Extensive experiments under various types of state-of-the-art attacks and diverse attack scenarios demonstrate the advantages of X-Ensemble to competitive baseline methods.
\end{abstract}


%

\maketitle
\section{Introduction}

Recent years have witnessed the unprecedented success of neural network-based deep learning (DL) models being widely deployed in many critical applications, such as financial investment and trading~\cite{wang2019alphastock}, and route recommendation~\cite{
guo2019rod,
wang2019empowering}. But with power comes a fatal weakness: the fragility of DL models when they face adversarial attacks. As early proposed by Szegedy \etal~\cite{2013szegedy}, it is very easy to mislead the outputs of a DL model by slightly perturbing input examples to form adversarial examples. This weakness causes DL models very unreliable in applications that are vulnerable to deliberate attacks, such as dangerous goods transportation ~\cite{wang2017no} and military systems, and eventually could lead to substantial economic losses and even the costs of lives.

In the literature, tremendous research efforts have been devoted to designing adversary defense (AD) methods to mitigate the threats of adversarial examples. Studies in this stream can be roughly divided into two categories: one is to strengthen target models~\cite{madry2017pgd,rony2019ddn} and the other is to detect and drop adversarial examples~\cite{song2017pixeldefend,hu2019new,lee2018simple}. While these methods have made great success in adversary defense, most of them yet suffer from three fundamental and long-standing challenges. The first challenge is to explore the intrinsic mechanism of adversarial attacks so as to enhance the defense ability of DL methods. The second challenge is to defense hybrid adversarial attacks that might include various types of attacks or even unknown types. The third challenge is an interesting game problem: how to protect a defender itself from adversarial attacks? These challenges indeed motivate our study in this paper.

We argue that the adversarial attacks have a close connection to another widely-admitted fatal weakness of DL models: the unexplainable problem~\cite{guidotti2019survey}. Deep neural networks are typically regarded as ``black-boxes'' since their stacked model structures are very hard to understand by human intuitions, which as the attack vulnerability also makes them not trustworthy in real-life critical applications. While there has been a lot of research on improving the interpretability of DL models, it is not until quite recently there appear some works that consider both the interpretation and the adversarial attacks from a cross-cutting perspective. It is reported that some types of local interpretations are more sensitive to adversarial attacks~\cite{gu2019saliency} and a more robust neural network has better interpretability~\cite{etmann2019connection}. Along this line, we may find the Achilles' heel of adversarial examples and design interpretation-based models for high-performance adversary defense.

In this work, we address the above three AD challenges by designing an interpreter-based ensemble framework called \name for robust adversary defense. Specifically, we reveal a gradient-based correlation between the sensitivity analysis-based interpreters and the generation process of adversarial examples, which implies that the Achilles' heel of an adversarial example is the interpretation map. Following this line, we propose the \name model, which adopts a detection-rectification pipeline and consists of three main components: $i$) A group of interpreter-based sub-detectors to classify input samples as adversarial/benign ones. The sub-detectors are all DNN-based models that can achieve high performance through fully exploiting different types of interpretation information toward target classifiers. $ii$) An ensemble mechanism uses a Random Forest model to combine sub-detector as a final detector. On one hand, an ensemble can improve the robustness of sub-detectors for hybrid attacks detection. On the other, the non-differentiable property of Random Forest (RF) can help against adversarial attacks toward detectors themselves. $iii$) An interpreter-based rectifier recovers adversarial examples to benign ones. With the guidance of model interpreters, our rectifier can purposefully erase pixels perturbed by attackers in a randomized mechanism.

Our main contribution can be summarized as follows:

$\bullet$ First, we reveal an underlying connection between sensitivity analysis-based interpretation of DL models and adversarial examples generation. We leverage this fundamental connection to design the detector and rectifier of \name, and all achieved advantageous performances (\ie solution to the first AD challenge).

$\bullet$ Second, we propose an RF-based framework to combine multiple types of interpreters as an ensemble detector for adversarial attacks of various types (\ie solution to the second AD challenge). The non-differentiable property of RF makes our detector much more robust compared with existing differentiable detectors (\ie solution to the third AD challenge).

$\bullet$ Finally, the performance of \name is verified over three popular image data sets under the attacks of five types of state-of-the-art adversarial examples. For diversified attack scenarios, including white-/black-/grey-box threat models,  vaccinated/unvaccinated training, as well as targeted/untargeted adversarial specificities, our model all shows superior accuracy and robustness.




\section{The Model Framework}
\label{sec:framework}

Threats of adversarial examples exist in many applications, such as image classification~\cite{resnet} and natural language processing~\cite{zhang2019adversarial}, and traffic speed prediction~\cite{wang2016traffic}. In this work, we select image classification as an example to introduce our model, which can be extended to other application scenarios.

We use a matrix $\bm{X} \in \mathbb{R}^{I\times J}$ to denote an image with $I\times J$ pixels.
Given an image $\bm{X}$, we have a one-hot coded label, denoted as $\bm{y}^\circ \in \{0,1\}^L$, to indicate its category. Note that $\|\bm{y}^\circ\|_1 = 1$, \ie only one item equals 1 in $\bm{y}^\circ$, and the $l$-th item equaling 1 means the image belongs to the class $l$, $l=1,\cdots,L$. For adversarial examples in image classification, there exists a target classifier and an attacker.

\begin{myDef}[\bf Classifier]
Given a set of images, we train a {\em classifier} ${F}: \mathbb{R}^{I\times J} \rightarrow [0,1]^L$. For each image sample $\bm{X}$ with a true label $\bm{y}^\circ$, the predicted label given by $F$ is denoted as $\hat{\bm{y}} = {F}(\bm{X})$. We assume ${F}$ is well trained and $\hat{\bm{y}} = \bm{y}^\circ$ for most of samples.
\end{myDef}
In this work, we assume the classifier is implemented using deep neural network models. 

\begin{myDef}[\bf Attacker]
Given a classifier $F$ for a set of images, we train an {\em attacker} ${A}: \mathbb{R}^{I\times J} \rightarrow \mathbb{R}^{I\times J}$. For each original image $\bm{X}^\circ$ with a true label $\bm{y}^\circ$, the attacker can generate an adversarial counterpart $\bm{X}^{(a)} = {A}(\bm{X}^\circ)$, satisfying that ${F}(\bm{X}^{(a)}) \neq \bm{y}^\circ = {F}(\bm{X}^\circ)$ and $\mathrm{Dist}(\bm{X}^\circ, \bm{X}^{(a)}) < \epsilon$, where $\mathrm{Dist}(\cdot, \cdot)$ is a distance measurement between two images, and $\epsilon$ is a small threshold to limit the difference between a benign example and its adversarial counterpart.
\end{myDef}

The distance function $\mathrm{Dist}(\cdot, \cdot)$ for attackers could be $l_0$, $l_2$, $l_\infty$ norms or others~\cite{yuan2019adversarial}. For the attacker $A$, the classifier $F$ is also called a target classifier for correspondence.




Since the classifier $F$ is a DNN-based ``black-box'' model, we use sensitivity-analysis-based interpreters~\cite{sundararajan2017axiomatic} to explain $F$. As will be given in Sec.~\ref{sec:connection}, sensitivity-analysis-based interpreters have close connections to adversarial examples, which are defined as follows.

\begin{myDef}[\bf Interpreter]
Given a classifier ${F}$, we define its interpreter as ${Ex}: \mathbb{R}^{I\times J} \rightarrow \mathbb{R}^{I\times J \times L}$. For an input image $\bm{X}$, the interpreter generates a sensitivity tensor $\mathcal{\bm{R}} = {Ex}(\bm{X})$, and the $(i,j,l)$-th item of $\mathcal{\bm{R}}$, \ie $r_{ijl}$, indicates the importance of $x_{ij}$ for $F$ if labeling $\bm{X}$ with the class $l$.
\end{myDef}

In this work, we propose a framework consisting of a detector and a rectifier for adversary defense, where the detector is in charge of detecting adversarial examples and the rectifier is in charge of recovering adversarial examples to benign ones. Below are the formal definitions of the two components.

\begin{myDef}[\bf Detector]
A detector is a function in the form of $Det:\mathbb{R}^{I\times J}\times\mathbb{R}^{I\times J\times L} \rightarrow \{0,1\}$. For an input image $\bm{X}$ and its classifier $F$, a detector gives a binary output
\begin{equation}\label{eq:detector}\small
z = Det(\bm{X},Ex(\bm{X})) \in \{0,1\},
\end{equation}
indicting that $\bm{X}$ is an adversarial example if $z=1$ and a benign example otherwise.
\end{myDef}

As reported by some previous works~\cite{carlini2017adversarial}
, using only $\bm{X}$ as feature to detect adversarial examples is not very robust. We therefore incorporate interpretations of $\bm{X}$ {\it w.r.t.} the target classifier into the detector, \ie $Ex(\bm{X})$, as given in the above definition. Similarly, we also include interpretations to build a rectifier as follows.

\begin{myDef}[\bf Rectifier]
A rectifier is a function in the form of $Rec:\mathbb{R}^{I\times J}\times\mathbb{R}^{I\times J\times L} \rightarrow \mathbb{R}^{I\times J}$. For an adversarial example $\bm{X}^{(a)}$ and its classifier $F$, a rectifier can convert it to a benign example as
\begin{equation}\small
\bm{X}^{(r)} = Rec(\bm{X}^{(a)},Ex(\bm{X}^{(a)})),
\end{equation}
such that $F(\bm{X}^{(r)}) = \bm{y}^{(r)} = \bm{y}^\circ$.
\end{myDef}

\begin{figure}
  \centering
  \includegraphics[width=0.9\columnwidth]{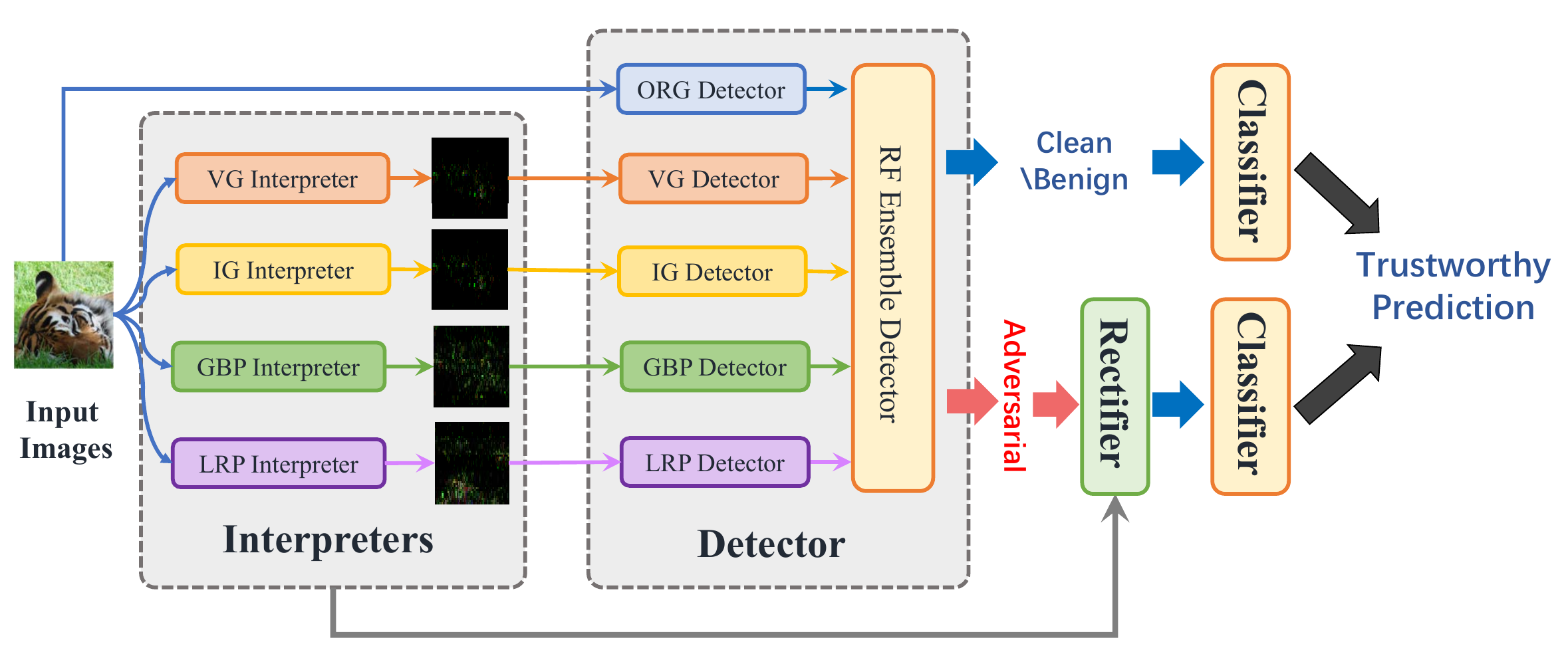}\\
  \caption{The framework of \name.}\label{fig:framework}
\end{figure}

\paratitle{Framework of \name.} Given the above definitions, we now introduce the general framework of our adversary defense model: \name, as shown in Fig.~\ref{fig:framework}, by combining the detector and the rectifier as a pipeline. Given an input image $\bm{X}$, the model first uses the detector $Det$ to classify it as a benign or adversarial example. If $\bm{X}$ is benign, \name uses the classifier $F$ to predict a label, otherwise, \name uses the rectifier to recover $\bm{X}$ as $\bm{X}^{(r)}$ and then applies $F$ on $\bm{X}^{(r)}$ to make the prediction. The image classification in our model is essentially in the form of
\begin{equation}\label{eq:H_combine}\small
  \bm{y} = \left\{\begin{aligned}
  & F\big(\bm{X}\big) &~~~ \mathrm{if}~~~ Det(\bm{X},\mathcal{\bm{R}}) = 0 \\
  & F\big(Rec(\bm{X},\mathcal{\bm{R}})\big) &~~~~ \mathrm{if}~~ Det(\bm{X},\mathcal{\bm{R}}) = 1
\end{aligned}\right. ,
\end{equation}
where $\mathcal{\bm{R}} = Ex(\bm{X})$. Fig.~\ref{fig:framework} is the illustration of Eq.~\eqref{eq:H_combine}, and more details about model training will be given in the following sections.

It is noteworthy that \name is novel in integrating the detection and rectification of adversarial examples into a unified process for high-quality prediction. \name is also novel in exploiting example-level interpretation information for building robust detectors and rectifiers, which requires a local interpreter for the classifier to be built in advance. 


\section{The Detector}

\subsection{The Idea of Detector}
\label{sec:connection}

As shown in Eq.~\eqref{eq:detector}, the detector uses the outputs of an interpreter as inputs. This design is motivated by the connection between adversarial attacks and model interpretations~\cite{etmann2019connection,gu2019saliency}, with the details introduced below.

The interpreters adopted by \name are gradient-based sensitivity analysis methods. The idea of this kind of interpreters is to describe the influences of inputs to outputs in a classifier. A simple method to describe this influences is using the gradient of the output $y^{(l)}=F_l(\bm{X})$ to the input pixel $x_{ij}$, \ie
\begin{equation}\label{eq:gijl}\small
g_{ijl} = \frac{\partial F_l(\bm{X})}{\partial x_{ij}},
\end{equation}
where $F_l$ is the $l$-th component of the classifier $F$'s output logits. A large value of $\left|~g_{ijl}~\right|$ means that the pixel $x_{ij}$ is an important factor for classifying the image $\bm{X}$ as the class $l$.

The gradient $g_{ijl}$ also plays an important role in adversarial examples generation. Given a raw image $\bm{X}^\circ$ and a target classifier $F$, the objective function of generating an adversarial example $\bm{X}^{(a)}$ that can fool $F$ to give an output $l^{(a)}$ is
\begin{equation}\label{eq:attacker_obj}\small
\begin{aligned}
&&\mathop{\arg\min}_{\bm{X}^{(a)}}~~\mathcal{L} \left(F\left(\bm{X}^{(a)}\right), l^{(a)}\right),\\
&&s.t.~ \mathrm{Dist}\left(\bm{X}^{(a)}, \bm{X}^\circ\right) < \epsilon,
\end{aligned}
\end{equation}
where $\mathcal{L}$ is a loss function and the cross entropy is the usual choice.

If we use the gradient descent method to optimize the objective function Eq.~\eqref{eq:attacker_obj}, the pixel $x_{ij}$ can be iteratively perturbed as
\begin{equation}\label{eq:attacker_update}\small
{x}_{ij}^{(\tau+1)} := \Gamma_{D_\epsilon(\bm{X}^\circ)}\underbrace{\left({x}_{ij}^{(\tau)} - \alpha \frac{\partial  \mathcal{L}\left(F\left(\bm{X}^{(\tau)}\right),~~ l^{(a)}\right)}{\partial {x}_{ij}^{(\tau)}}\right)}_{\mathrm{Term~I}},
\end{equation}
where $\Gamma_{D_\varepsilon(\bm{X}^\circ)}$ is the projection operator to ensure  $\mathrm{Dist}(\bm{X}^{(a)},\bm{X}^\circ) < \varepsilon$, $\tau$ is an iteration round index, and $\alpha$ is a learning rate. Using the chain rule, we rewrite {\em Term I} in Eq.~\eqref{eq:attacker_update} as
\begin{equation}\label{eq:attacker_update_2}\small
{x}_{ij}^{(\tau)} - \alpha \frac{\partial  \mathcal{L}}{\partial F_{l^{(a)}}\left(x_{ij}^{(\tau)}\right)}\cdot g_{ijl^{(a)}}.
\end{equation}
As shown in Eq.~\eqref{eq:attacker_update_2}, pixels with larger $\left|~g_{ijl^{(a)}}~\right|$ tend to be more seriously perturbed by attackers. Although different attackers may adopt different objective functions and optimization algorithms, the endogenous connection between adversarial attacks and $g_{ijl}$-based interpretations can be described schematically by Eq.~\eqref{eq:gijl}-Eq.~\eqref{eq:attacker_update_2}.

Inspired by the above connection between attackers and interpreters, we argue that the $g_{ijl}$ information can be adopted to detect adversarial examples. Specifically,

$\bullet$ {\em From the attacker perspective}, a pixel with large $|g_{ijl}|$ in an adversarial example tends to be perturbed until its $|g_{ijl}|$ is low enough and significantly different from its benign counterpart. In other words, although the attacker in Eq.~\eqref{eq:attacker_update} constrains the distance between $\bm{X}^\circ$ and $\bm{X}^{(a)}$ using the $\Gamma$ operator, the distance between the gradient matrices, \ie $\bm{G}_l^\circ$ and $\bm{G}_l^{(a)}$ consisting of $g_{ijl}$, is not constrained consciously and might provide important clues for adversarial and benign examples differentiation.

$\bullet$ {\em From the interpreter perspective}, the interpretation matrix $\bm{G}_l$ indicates the pixels that have decisive influences to image classification of the label $l$. For a benign image with a correct classification, the distribution of high $|g_{ijl}|$ pixels should be in line with human's visual cognition, \ie heavily distributed on indication objects of an image, such as a dog in an image. However, this might not be the case for an adversarial image with a misleading classification. In other words, we could expect a great difference for high $|g_{ijl}|$ pixel distributions in a benign example and its adversarial counterpart. Therefore, we can train a model to classify benign and adversarial examples by leveraging this difference.

The above insights indeed motivates the design of our detector for the \name model below.

\subsection{Interpreters in \name}
\label{sec:interpretation}

Given the effectiveness of interpretation information for detector training, the detector is yet vulnerable to adversarial attacks if only one kind of interpreters is used. Thus, we adopt an ensemble approach to combine multiple interpretation methods as the detector of \name. The interpretation methods adopted by \name include four types: Vanilla Gradient (VG), Integrated Gradient (IG)~\cite{sundararajan2017axiomatic}, Guided Backpropagation (GBP)~\cite{springenberg2014striving} and Layer-wise Relevance Propagation (LRP)~\cite{montavon2019layer}. The principle of method selection is to allow as much diversity as possible.

\paratitle{Vanilla Gradient.} Given a classifier $\hat{\bm{y}} = F(\bm{X})$, the influence of pixels in $\bm{X}$ to the one-hot classification output $\hat{\bm{y}}$ in the vanilla gradient is calculated as
\begin{equation}\label{eq:GRD}\small
    \mathcal{G}^{(VG)}=\frac{\partial F{(\bm{X})}}{\partial \bm{X}},
\end{equation}
where $\mathcal{\bm{G}} \in \mathbb{R}^{I\times J\times L}$ is the gradient matrix {\it w.r.t.} $F$. The $(i,j,l)$-th element of $\mathcal{G}$, \ie $g_{ijl}$, expresses the influence of pixel $x_{ij}$ to the class $l$.

\paratitle{Integrated Gradient.} The integrated gradient calculates the influence of pixels ${x}_ij$ to output ${y}_l$ as an integration of vanilla gradient~\cite{sundararajan2017axiomatic}, \ie
\begin{equation}\label{}\small
{{g}_{ijl}}^{(IG)} = ({x}_{ij}-{x}'_{ij})\times \int_{\alpha=0}^{1} \frac{\partial F_l\left({x}_{ij}'+\alpha({x}_{ij}-{x}_{ij}')\right)}{\partial {x}_{ij}}\, \mathrm{d}\alpha,
\end{equation}
where ${x}_{ij}'$ is the pixel of a baseline image usually set as a blank picture. The tensor $\mathcal{\bm{G}}^{(IG)}$, which consists of $g_{ijl}$, is proposed to avoid the gradient saturation problem of the vanilla gradient for neural network interpretation.

\paratitle{Guided Backpropagation.} GBP
modifies the local gradient of ReLU activation neuron~\cite{springenberg2014striving}. For a neural network classifier $F$, we denote $f_{m}$ as the output of a ReLU activation neuron at the $m$-th layer, and $r_m$, $r_{m+1}$ are the vanilla gradients of $\hat{\bm{y}}$ to the ReLU neuron's output and input respectively in gradient backward pass chain. GBP modifies the output gradient of a ReLU activation neuron as
\begin{equation}\label{eq:gbp}\small
    r_m=\mathds{1}(f_{m-1}>0) \cdot \mathds{1}(r_{m+1}>0) \cdot r_{m+1},
\end{equation}
where $\mathds{1}(\cdot)\in \{0,1\}$ is an indicative function. GBP uses Eq.~\eqref{eq:gbp} to calculate local gradient of each ReLU neuron, and uses the vanilla gradient defined in Eq.~\eqref{eq:GRD} to calculate the end-to-end interpretation tensor as $\mathcal{\bm{G}}^{(GBP)}$. The GBP method is proposed to improve the interpretation performance of gradient methods for neural networks with ReLU activation functions.

\paratitle{Layer-wise Relevance Propagation.} LRP is an approximate method of Taylor decomposition~\cite{montavon2019layer}. LRP assigns each pixel $x_{ij}$ a relevance score $g_{ijl}$, which have a relation with $y_l$ as
\begin{equation}\label{eq:lrp}\small
  y_l = \sum_{ij} g_{ijl}.
\end{equation}
The $g_{ijl}$ is calculated along the backward pass of neural networks layer by layer. Specifically, for the $m+1 \rightarrow m$ layers, LRP calculates a relevance $r_{m}^{(p)}$ for the $p$-th neuron in the layer $m$ as
\begin{equation}\label{}\small
  r_{m}^{(p)} = \sum_q r_{m+1}^{(q)}\frac{f_{m+1}^{(p)} w_{pq}}{\sum_h f_{m+1}^{(h)} w_{hq}}
\end{equation}
where $f_{m+1}^{(p)}$ is the output of the $p$-th neuron in the layer $m+1$, and $w_{pq}$ is the weight connecting the neurons $f_{m+1}^{(p)}$ and $f_{m}^{(q)}$. LRP is considered as a type of gradient-based interpretation method.
The tensor composed of $g_{ijl}$, denoted as $\mathcal{\bm{G}}^{(LRP)}$, is adopted as the input of our detector.

\subsection{Ensemble Detector}

The ensemble detector of \name consists of multiple sub-detectors and an ensemble classifier.

\subsubsection{Sub-Detectors}

For each interpreter, \name trains a Convolutional Neural Network (CNN) to classify image $\bm{X}$ as an adversarial or benign example. The CNN's input is $\bm{G}_l$ generated by the interpreter, where $l$ is the index of the class predicted by the original image classifier $F$ and hence $\bm{G}_l$ is the slice in the tensor $\mathcal{\bm{G}}$ corresponding to class $l$. We call these adversarial/benign example classifiers as sub-detectors of \name. For the four interpreters in Sec.~\ref{sec:interpretation}, the corresponding sub-detectors are in the form of
\begin{equation}\label{eq:detector_inter}\small
\begin{aligned}
  &\bm{z}^{(VG)} = \mathrm{CNN}_1\left(\bm{G}^{(VG)}_l\right),&~&\bm{z}^{(IG)} = \mathrm{CNN}_2\left(\bm{G}^{(IG)}_l\right),\\
  &\bm{z}^{(GBP)} = \mathrm{CNN}_3\left(\bm{G}^{(GBP)}_l\right),&~&\bm{z}^{(LRP)} = \mathrm{CNN}_4\left(\bm{G}^{(LRP)}_l\right).
\end{aligned}
\end{equation}
Besides the four interpretation-based sub-detectors, we also adopt original images as inputs to train a sub-detector as
\begin{equation}\label{eq:detector_org}\small
    \bm{z}^{(ORG)} = \mathrm{CNN}_5(\bm{X}).
\end{equation}

To train the above sub-detectors, we need to adopt attack algorithms to prepare adversarial examples. In the literature, however, there exists a large number of attackers~\cite{yuan2019adversarial}, and it is unpractical to train a set of sub-detectors for each attacker. We therefore group attackers according to their perturbation measurements, which typically include $l_0$, ${l}_2$, and $l_\infty$. To further simplify the problem, we only adopt $l_2$ and $l_\infty$ measurements in our experiments because they are the most widely used in the state-of-the-art attackers. In fact, the performance of using $l_2$ and $l_\infty$ measurements to detect $l_0$ attackers is also satisfactory enough, which we will show in Sec.~\ref{sec:experiment}.

Given a clean example set $\mathcal{X}=\{\bm{X}_1, \ldots, \bm{X}_N\}$, we use different adversarial models to generate two sets of adversarial counterparts: one is generated by $l_2$ attackers and the other is generated by $l_\infty$ attackers. We mix $\mathcal{X}$ and the two sets of adversarial examples as a training set. For a training example, we set a label $\hat{\bm{z}} \in \{0,1\}^3$, where $\hat{\bm{z}} = (1,0,0)$ indicates clean example, $\hat{\bm{z}} = (0,1,0)$ indicates $l_2$ attack, and $\hat{\bm{z}} = (0,0,1)$ indicates $l_\infty$ attack. Finally, we use the training set to train the sub-detectors defined in Eq.~\eqref{eq:detector_inter} and Eq.~\eqref{eq:detector_org}.

\subsubsection{Ensemble Classifier}

\name adopts a random forest (RF) to combine the sub-detectors as an ensemble adversarial/benign classifier, \ie
\begin{equation}\label{eq:detector_ensemble}\small
{z} = RF\left(\bm{z}^{(ORG)}, \bm{z}^{(VG)}, \bm{z}^{(IG)}, \bm{z}^{(GBP)}, \bm{z}^{(LRP)}\right).
\end{equation}
The ensemble detector no longer classify adversarial examples by their perturbation measurements. The output $z$ is the predicted benign/adversarial label, \ie ${z} = 0$ for benign examples and ${z} = 1$ for adversarial examples.

Compared with CNN-based models, the random forest model is non-differentiable and therefore not easy to be directly attacked. Moreover, the random forest as an ensemble detector is much more robust than a single sub-detector when facing different types of adversarial attacks even unknown ones.


\begin{table}[t!]
  \caption{Attack success rate with largest $g_{ijl}$ pixels dropped.}
  \centering
  \footnotesize
  \label{tab:toyexp}
    \begin{tabular}{c|ccc}
    \toprule
    {Erased Rate}     & Deepfool & CW & DDN \\\midrule
    top 0\% & 1.000  & 1.000  & 1.000  \\
    top 5\% & 0.637  & 0.665  & 0.656  \\
    \bottomrule
    \end{tabular}
	\vspace{-0.2cm}
\end{table}%

\section{Rectifier}

Most of traditional studies in adversarial attack detection directly drop suspicious examples to protect target classifiers~\cite{hu2019new,lee2018simple}. Here, we go beyond the detect-and-drop scheme and set a rectifier to recover adversarial examples as benign ones.

The proposed rectifier also has a close connection with interpreters. In a sensitivity-based interpretation map $\bm{G}_l$, a large value of $g_{ijl}$ indicates the pixel $x_{ij}$ is an important factor for assigning the image $\bm{X}$ with the class $l$. For an adversarial example, its predicted label $l$ is unreliable, so
the large $g_{ijl}$ pixels
are very likely to be modified by attackers to mislead the classifier.
Therefore, it is intuitive to erase the pixels with large $g_{ijl}$ in an adversarial example to weaken the attack effects.

Table~\ref{tab:toyexp} is a toy experiment. We use three attackers, \ie Deepfool~\cite{moosavi2016deepfool}, CW~\cite{carlini2017cw} and DDN~\cite{rony2019ddn}, to generate adversarial examples for a classifier model, and set the pixels with top 5\% largest $g_{ijl}$ in the adversarial examples as zero, where $g_{ijl}$ is the vallina gradient. As can be seen from the table,
the success rates of the original adversarial examples are 100\% while the modified ones fall to only 63.7\%.
This confirms our argument that $g_{ijl}$ can be used to pick polluted pixels out from adversarial examples. We design our rectifier based on this observation accompanied by a random erasion scheme.

Given an adversarial example $\bm{X}^{(a)}$ with its predicted label $l$, our rectifier selects a pixel as a suspect when its $g_{ijl}$ is larger than a threshold ; that is,
\begin{equation}\label{eq:masking_rate}\small
    g_{ijl} > \alpha \cdot \Big(g_{max} - g_{min}\Big) + g_{min},
\end{equation}
where $g_{max}$ and $g_{min}$ are the maximum and minimum values among all pixel's $g_{ijl}$ in the adversarial example, and $\alpha$ is a preset parameter to control the ratio of suspect pixels.

For a suspect pixel $x_{ij}$, the rectifier randomly erases it by setting
\begin{equation}\label{eq:random}\small
    x^{(r)}_{ij} : = \left\{\begin{aligned}
  & x^{(a)}_{ij} &~~~ \mathrm{if}~~ u_{ran} = 0, \\
  & x^{(a)}_{ij} + x_{ran} &~~~ \mathrm{if}~~ u_{ran} = 1,
\end{aligned}\right.
\end{equation}
where $u_{ran} \in \{0,1\}$ follows a Bernoulli distribution with $p=0.5$ and controls whether a pixel needs to be erased.
If $u_{ran} = 0$, we set $x^{(r)}_{ij}$ as its raw value, otherwise we add it with a random value $x_{ran}$ that follows a normal distribution $N(0, \sigma)$, with $\sigma$ being the standard deviation of pixels in $\bm{X}^{(a)}$.

The random erasure mechanism of the rectifier enables us to generate multiple randomized duplicates for every adversarial example,
to fine-tune the original classifier $F$
and improve its ability to handle the rectified examples. Finally, we use the fine-tuned classifier to predict the labels of rectified adversarial examples.

So far, there remains one issue in the implementation of the rectifier: how to calculate $g_{ijl}$ given the four types of interpreters in the ensemble detector. We follow two principles to solve this: $i$) The chosen interpreter-based sub-detector should have the same detection result with the ensemble detector; $ii$) Given an input example $\bm{X}$, the chosen interpreter-based sub-detector should have the lowest information entropy on its prediction $\hat{\bm{z}} = (z_1, z_2, z_3),~\sum_i z_i = 1$, which is calculated as
\begin{equation}
  H^{(pred)} = - \sum_{i=1}^3 z_i \log_2(z_i).
\end{equation}
A lower $H^{(pred)}$ means the information provided is richer than others, which implies that the $g_{ijl}$ adopted in Eq.~\eqref{eq:masking_rate} can offer more information to adversarial examples rectification.

\section{Experiments}
\label{sec:experiment}

\subsection{Datasets and Attackers}

In the experiments, we employ three popular image datasets, namely Fashion-MNIST~\cite{xiao2017fashion}, CIFAR-10~\cite{krizhevsky2009cifar} and ImageNet~\cite{imagenet_cvpr09}.

The experiments are conducted on five extensively used and state-of-the-art attack algorithms. They are {\em $l_\infty$ attackers}: FGSM~\cite{fgsm}, PGD~\cite{madry2017pgd}, and {\em $l_2$ attackers}: Deepfool (DFool)~\cite{moosavi2016deepfool}, CW~\cite{carlini2017cw}, DDN~\cite{rony2019ddn}.

These attacks have two types of adversarial specificities, \ie untargeted attacks and targeted attacks~\cite{akhtar2018threat}. The targeted attacks mislead the output of a classifier to a given category, while untargeted attacks mislead the output to any false category. In the experiments, we mark the untargeted version of attackers with the postfix ``-U'', and  the targeted with ``-T''. FGSM and Deepfool only have untargeted versions, and others have both. We finally have eight types of different attackers.

Attackers in our experiments have three threat models:

$\bullet$ {\em Grey-box.} Attackers have knowledge about the network parameters and structure of the classifier but are not aware that the classifier is under the protection of any defence, so they generate adversarial examples that can only fool the target classifier.

$\bullet$ {\em Black-box.} Attackers have no access to the model of the target classifier, and are not aware that the classifier is under protections. They have to use adversarial examples generated for a surrogate classifier to attack the target classifier.

$\bullet$ {\em White-box.} Attackers have full knowledge about the target classifier and the defence scheme, so they can generate adversarial examples to fool the entire \name model.


\begin{table*}[t!]\footnotesize
  \centering
  \caption{Adversarial examples detection performance for vaccinated training.}
  \vspace{-0.2cm}
  \resizebox{0.95\textwidth}{!}{
    \begin{tabular}{c|c|c|c|c|c|c|c|c|c|c|c|c|c|c|c|c|c|c}
    \toprule
    \multicolumn{19}{c}{Grey-Box} \\
    \midrule
      & \multicolumn{9}{c|}{Fashion-MNIST} & \multicolumn{9}{c}{CIFAR10} \\
    \midrule
    Attackers & X-Det & PD & TWS  & MDS  & VG & IG & GBP & LRP & ORG & X-Det & PD &  TWS & MDS  & VG & IG & GBP & LRP & ORG  \\
    \midrule
        FGSM-U  & \textbf{1.00 } & \textbf{1.00 } & 0.63  & 0.71  & 0.97  & 0.99  & \textbf{1.00 } & 0.99  & \textbf{1.00 } & \textbf{1.00 } & 0.98  & 0.52  & 0.83  & 0.88  & 0.86  & 0.98  & 0.99  & \textbf{1.00 } \\
        PGD-U  & \textbf{1.00 } & \textbf{1.00 } & 0.65  & 0.79  & 0.98  & \textbf{1.00 } & 0.99  & 0.99  & \textbf{1.00 } & \textbf{0.99 } & \textbf{0.99}  & 0.52  & 0.76  & \textbf{0.99 } & 0.95  & 0.96  & 0.97  & 0.98  \\
        PGD-T  & \textbf{1.00 } & \textbf{1.00 } & 0.83  & 0.80  & 0.97  & \textbf{1.00 } & 0.99  & 0.99  & \textbf{1.00 } & 0.98  & 0.96  & 0.48  & 0.71  & 0.93  & 0.90  & 0.95  & 0.98  & \textbf{1.00 } \\\hline
        DFool-U  & 0.99  & 0.98  & 0.99  & 0.77  & 0.95  & 0.99  & \textbf{1.00 } & 0.94  & 0.99  & 0.98  & 0.77  & 0.83  & 0.93  & 0.89  & 0.90  & \textbf{0.99 } & 0.92  & 0.83  \\
        CW-U  & 0.98  & 0.93  & 0.95  & 0.79  & 0.94  & 0.98  & \textbf{1.00 } & 0.98  & 0.96  & 0.98  & 0.78  & 0.90  & 0.93  & 0.90  & 0.89  & \textbf{0.99 } & 0.92  & 0.86  \\
        CW-T  & \textbf{1.00 } & 0.98  & 0.99  & 0.83  & 0.97  & \textbf{1.00}  & \textbf{1.00 } & \textbf{1.00 } & 0.99  & \textbf{0.99}  & 0.84  & 0.94  & 0.94  & 0.93  & 0.93  & \textbf{0.99 } & 0.96  & 0.95  \\
        DDN-U  & 0.99  & 0.98  & 0.80  & 0.79  & 0.96  & 0.99  & 0.99  & \textbf{1.00 } & 0.99  & \textbf{0.99}  & 0.70  & 0.91  & 0.93  & 0.91  & 0.90  & 0.92  & \textbf{0.99 } & 0.90  \\
        DDN-T  & \textbf{1.00 } & 0.99  & \textbf{1.00}  & 0.85  & \textbf{1.00 } & 0.90  & 0.98  & \textbf{1.00 } & \textbf{1.00 } & \textbf{0.99}  & 0.81  & 0.96  & 0.94  & \textbf{0.99 } & 0.93  & 0.95  & \textbf{0.99 } & 0.97  \\
    \toprule
    \multicolumn{19}{c}{Black-Box}  \\
    \midrule
      & \multicolumn{9}{c|}{Fashion-MNIST} & \multicolumn{9}{c}{CIFAR10}  \\
    \midrule
    Attackers & X-Det & PD & TWS  & MDS  & VG & IG & GBP & LRP & ORG & X-Det & PD &  TWS & MDS  & VG & IG & GBP & LRP & ORG  \\
    \midrule
        FGSM-U  & \textbf{1.00 } & 0.99  & 0.76  & 0.54  & \textbf{1.00 } & 0.98  & 0.99  & \textbf{1.00 } & \textbf{1.00 } & 0.98  & 0.99  & 0.66  & 0.93  & 0.88  & 0.92  & 0.99  & 0.99  & \textbf{1.00 } \\
        PGD-U  & \textbf{1.00 } & 0.99  & 0.77  & 0.53  & \textbf{1.00 } & 0.98  & 0.99  & \textbf{1.00 } & \textbf{1.00 } & 0.97  & 0.98  & 0.57  & 0.59  & 0.76  & 0.80  & 0.91  & 0.98  & \textbf{1.00 } \\
        PGD-T  & \textbf{1.00 } & 0.99  & 0.78  & 0.55  & \textbf{1.00 } & 0.97  & 0.99  & \textbf{1.00 } & \textbf{1.00 } & 0.99  & 0.99  & 0.72  & 0.59  & 0.78  & 0.83  & 0.92  & 0.96  & \textbf{1.00 } \\\hline
        DFool-U  & 0.94  & 0.93  & 0.81  & 0.52  & 0.85  & 0.94  & \textbf{0.98 } & 0.91  & 0.95  & 0.79  & 0.74  & 0.75  & 0.54  & 0.70  & \textbf{0.80 } & \textbf{0.80 } & \textbf{0.80 } & 0.60  \\
        CW-U  & 0.91  & 0.87  & 0.81  & 0.53  & 0.83  & 0.91  & \textbf{0.99 } & 0.90  & 0.86  & \textbf{0.82}  & 0.75  & 0.75  & 0.53  & 0.71  & \textbf{0.82 } & 0.80  & 0.81 & 0.70  \\
        CW-T  & 0.97  & 0.96  & 0.80  & 0.52  & 0.91  & \textbf{0.99}  & {0.98 } & 0.95  & 0.98  & \textbf{0.82 } & 0.77  & 0.76  & 0.53  & 0.80  & \textbf{0.82 } & \textbf{0.82 } & \textbf{0.82 } & 0.77  \\
        DDN-U  & 0.88  & 0.86  & 0.80  & 0.52  & 0.82  & \textbf{0.95 } & 0.94 & 0.91  & 0.93  & 0.80  & 0.63  & 0.76  & 0.54  & 0.71  & 0.80  & \textbf{0.81 } & 0.80  & 0.76  \\
        DDN-T  & 0.98  & 0.96  & 0.79  & 0.54  & 0.92  & 0.97  & \textbf{0.99 } & 0.96  & \textbf{0.99}  & 0.82  & 0.72  & 0.76  & 0.54  & 0.71  & 0.80  & 0.82  & 0.82  & \textbf{0.89 } \\
    \bottomrule
    \end{tabular}%
    }
  \label{tab:vaccinated}%
\end{table*}%

\subsection{Performance of Detector}

Here, we evaluate the performance of the detector in \name. We compare our ensemble detector, namely \dname, with three types of baselines using different feature engineering methods. They are:

$\bullet$ {\em PixelDefend (PD)}~\cite{song2017pixeldefend}. It utilizes a PixelCNN network to extract features of examples, and adopts a test statistic to decide whether an input is an adversarial example. This baseline is a representative of the detectors that use neural networks to extract features of adversarial examples.

$\bullet$ {\em TWS}~\cite{hu2019new}. It utilizes two seemingly paradoxical criteria of benign images to detect adversarial examples, with the assumption that an adversarial example cannot satisfy the two criteria. It could be considered as a type of hand-crafted feature-based detectors.

$\bullet$ {\em MDS}~\cite{lee2018simple}. It utilizes Mahalanobis Distance-based Score (MDS) to measure different distributions of adversarial and benign examples in a network's middle-layer outputs. This baseline is considered as a type of distribution feature-based detectors.

Since \name is an ensemble model, the sub-detectors are also given as baselines to evaluate the performance of our ensemble mechanism, named with their interpreters' names. We report the binary classification performance of the sub-detectors, \ie the outputs of both $l_2$ and $l_\infty$ are treated as adversarial examples. Therefore, we use AUC, \ie Area Under the Curve (AUC) of the Receiver Operating Characteristic (ROC), to measure detectors' performance.


In the experiments, we use different attackers to generate adversarial examples, and mix them with the original benign ones to train and test the detectors. Our experiments employ two kinds of training methods, namely {\em Vaccinated Training} and {\em Unvaccinated Training}. In the vaccinated training, adversarial examples in the training and test sets are generated by the same kind of attackers. Oppositely, in the unvaccinated training, adversarial examples are generated by different attackers. For example, if we train a detector using adversarial examples generated by the FGSM algorithm, for vaccinated training, we also need to use FGSM to attack our model for test (\ie our model has been ``vaccinated'' by FGSM ); but for unvaccinated training, we may use DDN or other attackers for test (\ie our model has not been ``vaccinated'' by unknown attackers).

We first give the experimental results of vaccinated training, where we use all eight attackers to generate adversarial examples to train our detector and then fool our model using the attackers one by one. The detection performance is tested on Fashion-MNIST and CIFAR-10 datasets. The AUC scores of the detectors under grey- and black-box attacks are given in Table~\ref{tab:vaccinated}. As can be seen,

$\bullet$ In general, neural network-based detectors, including PixelDefend, \dname and sub-detectors, have better performances than distribution and hand-crafted feature-based methods, indicating the neural network is more effective than traditional feature engineering approaches for extracting the features of adversarial examples.

$\bullet$ For the $l_\infty$ attacks, \ie FGSM and PGD, the performance of \dname is superior to TWS and MDS and is very close to PixelDefend. For the $l_2$ attackers, \ie Deepfool, CW and DDN, the performance of \dname is significantly better than TWS, MDS and PixelDefend. Even for the cases where \dname is not as better as PixelDefend, the AUC scores of \dname are more than 97\%, indicating pretty good performance. These verify the effectiveness of \dname.

$\bullet$ \name shows greater advantages on the CIFAR-10 dataset than on the Fashion-MNIST dataset. Since CIFAR-10 (color images) is deemed more complex than Fashion-MNIST (grayscale images), this implies that our model suits complex application scenarios.

$\bullet$ For most of the rows, \dname's performance is just a little worse than the best sub-detector. In fact, the primary task of the RF ensemble in \dname is not for improving the detection accuracy but for enhancing robustness. As the table shows, we cannot find a sub-detector that can always outperform the others, while \dname's performance is very stable across all cases, showing the precious ability to resist hybrid attacks.


$\bullet$ The detection performance under black-box attacks is worse than that under grey-box attacks. In the black-box attack scenario, the detectors are trained for protecting the target classifier $F$ but the attackers are not trained for attacking $F$. This mismatch causes the detection success rates of black-box underperform that of grey-box. However, since the attack success rates of black-box are also substantially lower than that of grey-box, the performance loss will not cause severe problems. In Table~\ref{tab:vaccinated}, the black-box performance is reported for successful adversarial examples (see SI for details). We can see \dname still outperforms PixelDefned, TWS and MDS.

\begin{table}[h]\footnotesize
  \centering
  \caption{AUC score of adversarial examples detection for unvaccinated training.}
  \vspace{-0.2cm}
    \begin{tabular}{l|c|c|c|c|c|c|c|c}
    \toprule
    \multicolumn{9}{c}{Grey-Box} \\
    \midrule
          & \multicolumn{4}{c|}{Fashion-MNIST} & \multicolumn{4}{c}{CIFAR-10} \\
    \midrule
    Attacker & X-Det & PD & $l_\infty$-D & $l_2$-D & X-Det & PD & $l_\infty$-D & $l_2$-D  \\    \midrule
    PGD-U & \textbf{1.00} & \textbf{1.00} & \textbf{1.00} & 0.90 & \textbf{1.00} & 0.99 & \textbf{1.00} & 0.39 \\
    PGD-T & \textbf{1.00} & \textbf{1.00} & 0.99 & 0.91 & \textbf{1.00} & 0.99 & \textbf{1.00} & 0.50 \\ \hline
    CW-U & 0.95 & 0.93 & 0.73 & \textbf{0.97} & \textbf{0.98} & 0.78 & 0.49 & 0.97 \\
    CW-T & 0.98 & 0.98 & 0.84 & \textbf{0.99} & \textbf{0.99} & 0.84 & 0.49 & 0.98 \\
    DDN-U & 0.99 & 0.98 & 0.80 & \textbf{1.00} & \textbf{0.99} & 0.70 & 0.49 & 0.98 \\
    DDN-T & \textbf{1.00} & \textbf{1.00} & 0.93 & \textbf{1.00} & \textbf{0.99} & 0.81 & 0.49 & 0.98  \\ \hline
    OnePixel & \textbf{0.82} & 0.61 & 0.59 & 0.75 & \textbf{0.83} & {0.81} & 0.51 & 0.77 \\
    \toprule
    \multicolumn{9}{c}{Black-Box} \\
    \midrule
          & \multicolumn{4}{c|}{Fashion-MNIST} & \multicolumn{4}{c}{CIFAR-10} \\
    \midrule
    Attacker & X-Det & PD & $l_\infty$-D & $l_2$-D & X-Det & PD & $l_\infty$-D & $l_2$-D  \\    \midrule
    PGD-U & {\bf 0.99} & {\bf 0.99} & 0.98 & 0.91 & 0.99 & 0.99 & \textbf{1.00} & 0.70 \\
    PGD-T & {\bf 0.99} & {\bf 0.99} & 0.98 & 0.92 & 0.99 & 0.99 & \textbf{1.00} & 0.78 \\ \hline
    CW-U & {\bf 0.87} & 0.85 & 0.51 & 0.73 & \textbf{0.80} & 0.75 & 0.48 & 0.77 \\
    CW-T & {\bf 0.97} & 0.93 & 0.78 & 0.88 & \textbf{0.80} & 0.77 & 0.49 & 0.76 \\
    DDN-U & 0.85 & {\bf 0.88} & 0.53 & 0.83 & \textbf{0.80} & 0.63 & 0.49 & 0.75 \\
    DDN-T & {\bf 0.95} & 0.98 & 0.84 & 0.90 & \textbf{0.82} & 0.72 & 0.48 & 0.77  \\
    \hline
    OnePixel & \textbf{0.73} & 0.71 & 0.57 & 0.69 & \textbf{0.72} & {0.70} & 0.51 & 0.69 \\
    \bottomrule
    \end{tabular}%
  \label{tab:unvaccinated}%
  \vspace{-0.4cm}
\end{table}%

Table~\ref{tab:unvaccinated} reports the performance of \name using {\em Unvaccinated Training}. Here, the training examples of \name are generated by two types of attackers, \ie FGSM as an $l_\infty$ attacker and DeepFool as an $l_2$ attacker. We use targeted and untargeted versions of PGD, CW and DDN attacks to evaluate the performance. 
The \dname model is compared with three baselines: $i)$ PixelDefend, which has the best performance among the baselines. $ii)$ $l_\infty$-D, which is a \dname model using only unvaccinated training by FGSM. $iii)$ $l_2$-D, which is a \dname model using only unvaccinated training by DeepFool. From Table~\ref{tab:unvaccinated}, we can observe that:

$\bullet$ $l_\infty$-D is effective for $l_\infty$ attackers, \ie PGD, but is invalid for $l_2$ attackers, \ie CW and DDN. This indicates that the difference between $l_\infty$ and $l_2$ adversarial examples is very significant, while the difference between $l_\infty$ (or $l_2$) adversarial examples is relatively small. The performance of \dname is better than that of both $l_\infty$-D and $l_2$-D, indicating the effectiveness of \dname that classifies examples into three classes ($l_\infty$, $l_2$ and clean) in sub-detectors.


$\bullet$ The last row of Table~\ref{tab:unvaccinated} gives the detection performance of $l_0$ adversarial examples generated by OnePixel attack~\cite{su2019one}. As can be seen, \dname achieves superior performances compared with the PD baseline. Note that in unvaccinated training, our model is not trained by $l_0$ adversarial examples, so this result further verifies the adaptability of \dname.


To conclude, the above experiments demonstrate the effectiveness of the RF  ensemble detector of \name.

\subsection{Robustness of Detector}
\label{sec:det_robust}

\begin{table}[t]\footnotesize
  \centering
  \caption{Robustness of detectors under attacks of transferable adversarial examples.}
    \begin{tabular}{c|c|c|c|c|c|c|c}
    \toprule
    \multicolumn{8}{c}{Fashion-MNIST} \\
    \midrule
    Targeted & \multicolumn{6}{c}{Detector Performance} & \multicolumn{1}{|c}{Success} \\
\cline{2-7}    \multicolumn{1}{c|}{Detector} &  ORG & VG & GBP & IG & LRP & X-Det & {Rate}\\
    \midrule
    toORG & {0.98} & 0.60 & 0.41 & 0.63 & 0.65 & \textbf{0.39} & 0.10  \\
    toVG & 0.59 & {0.64} & 0.34 & 0.58 & 0.57 & \textbf{0.33} & 0.11 \\
    toGBP & 0.63 & 0.55 &  {0.63} & 0.57 & 0.71 & \textbf{0.45} & 0.10 \\
    toIG & 0.66 & 0.69 & 0.37 & {0.98} & 0.93 & \textbf{0.37} & 0.10  \\
    \toprule
    \multicolumn{8}{c}{CIFAR10}\\
    \midrule
    Targeted & \multicolumn{6}{c}{Detector Performance} & \multicolumn{1}{|c}{Success} \\ \cline{2-7}
   {Detector} &  ORG & VG & GBP & IG & LRP & X-Det & {Rate} \\
    \midrule
    toORG & {0.98} & 0.56 & 0.53 & 0.55 & 0.57 & \textbf{0.34} & 0.11  \\
    toVG & 0.40 & {0.69} & 0.42 & 0.39 & 0.49 & \textbf{0.27} & 0.11 \\
    toGBP & 0.47 & 0.50 & {0.63} & 0.51 & 0.57 & \textbf{0.26} & 0.11 \\
    toIG & 0.38 & 0.60 & 0.60 & {0.98} & 0.63 & \textbf{0.28} & 0.10  \\
    \bottomrule
    \end{tabular}%
  \label{tab:robustness}%
  \vspace{-0.3cm}
\end{table}%

A type of threats to a detector is the white-box attack, \ie training a detector-targeted attacker to attack the detector directly. In fact, the RF ensemble of \dname is to defend detector-targeted attacks. To the best of our knowledge, there still exists no algorithms that can directly attack RF models. A feasible attack method is to use transferable adversarial examples~\cite{liu2016delving}. Following this way, we use transferable examples targeted to sub-detectors to attack \dname.

For each row of Table~\ref{tab:robustness}, the transferable adversarial examples are generated and targeted to one type of sub-detectors. For example, the first row is for the ORG sub-detector, denoted as ``toORG'', and we transfer the ``toORG'' examples to attack \dname and the sub-detectors. LRP is not differentiable in the second-order, so we cannot generate transferable examples from the LRP sub-detector. Table~\ref{tab:robustness} gives attack success rates of the transferable adversarial examples. The attack success rates are average of three types of attackers, \ie PGD-T, CW-T, DDN-T. The results show that:

$\bullet$ When the adversarial example generator and the test sub-detector are based on the same type of interpretation (see the diagonal line of the table), the attack success rates are significantly higher than others. Even so, the success rates of detector-targeted attacks are still significantly lower than that of classifier-targeted attacks. This suggests the interpreter-based sub-detectors are more robust than classifiers, although they are all neural network models.

$\bullet$ For most of cases, the attack success rate is in the range of 40\%-60\%. For some special cases, such as toGBP$\rightarrow$LRP and toIG$\rightarrow$LRP on Fashion-MNIST, it reaches 71\% and 93\%, respectively. This suggests that the transferable attacks indeed work for detector-targeted attacks, although the performance is mediocre.

$\bullet$ The attack success rates of transferable adversarial examples are less than 45\% to \dname. For all rows, \dname achieves the smallest success rates. This indicates the robustness of \dname compared with the sub-detectors and the effectiveness of the ensemble mechanism.

$\bullet$ As Table~\ref{tab:robustness} shows, there are still 26\%-45\% adversarial examples that escape the detection of \dname. In the last column of the table, we give the final attack success rates of the escaped adversarial examples to the classifier $F$, which are actually very low to about 10\%. This implies that while the transferable adversarial examples can threaten the \name's detector, it is very hard to fool both the detector and the classifier of \name as a whole.


\begin{table*}[t] \footnotesize
  \centering
  \caption{Image classification accuracy of \name and the baselines.}
  \resizebox{\textwidth}{!}{
    \begin{tabular}{c|c|c|c|c|c|c|c|c|c|c|c|c|c|c|c|c|c|c}
    \toprule
    \multicolumn{19}{c}{Grey-Box} \\
    \midrule
      & \multicolumn{6}{c|}{Fashion-MNIST} & \multicolumn{6}{c|}{CIFAR-10} & \multicolumn{6}{c}{ImageNet} \\
    \midrule
      & Our & PD & DDN$_a$ & PGD$_a$ & TVM & $F$ & Our & PD & DDN$_a$ & PGD$_a$ & TVM & $F$ & Our & PD & DDN$_a$ & PGD$_a$ & TVM & $F$ \\
    \midrule
    Clean & 0.90  & 0.90  & 0.86  & 0.84  & 0.67  & \textbf{0.92} & 0.82  & 0.79  & 0.75  & 0.64  & 0.35  & \textbf{0.86} & 0.89  & 0.66  & 0.78  & 0.72  & 0.75  & \textbf{0.95} \\ \hline
    FGSM-U  & \textbf{0.84} & 0.75  & 0.82  & 0.82  & 0.49  & 0.56  & \textbf{0.55} & 0.36  & 0.48  & 0.43  & 0.29  & 0.24  & \textbf{0.60} & 0.47  & 0.49  & 0.47  & 0.36  & 0.44 \\
    PGD-U & \textbf{0.79} & 0.64  & 0.80  & 0.81  & 0.57  & 0.27  & \textbf{0.41} & 0.30  & 0.37  & 0.35  & 0.32  & 0.08  & \textbf{0.75} & 0.70  & 0.38  & 0.47  & 0.66  & 0.02 \\
    PGD-T & \textbf{0.89} & 0.86  & 0.84  & 0.87  & 0.53  & 0.66  & \textbf{0.62} & 0.60  & 0.33  & 0.48  & 0.32  & 0.05  & \textbf{0.73} & 0.66  & 0.29  & 0.51  & 0.70  & 0.00 \\\hline
    Dfool-U & 0.87  & \textbf{0.88} & 0.26  & 0.76  & 0.65  & 0.00  & \textbf{0.71} & 0.68  & 0.19  & 0.29  & 0.34  & 0.00  & \textbf{0.75} & 0.58  & 0.37  & 0.35  & 0.71  & 0.01 \\
    CW-U  & 0.86  & \textbf{0.88} & 0.70  & 0.73  & 0.66  & 0.00  & \textbf{0.74} & 0.73  & 0.70  & 0.63  & 0.34  & 0.00  & \textbf{0.74} & 0.64  & 0.50  & 0.53  & 0.71  & 0.00 \\
    CW-T  & \textbf{0.86} & 0.85  & 0.72  & 0.53  & 0.65  & 0.00  & 0.74  & \textbf{0.75} & 0.45  & 0.46  & 0.33  & 0.00  & \textbf{0.79} & 0.61  & 0.40  & 0.39  & 0.75  & 0.00 \\
    DDN-U & \textbf{0.90} & 0.89  & 0.74  & 0.76  & 0.66  & 0.00  & 0.69  & \textbf{0.74} & 0.66  & 0.52  & 0.34  & 0.00  & \textbf{0.76} & 0.60  & 0.56  & 0.44  & 0.75  & 0.03 \\
    DDN-T & \textbf{0.90} & 0.89  & 0.59  & 0.64  & 0.65  & 0.00  & 0.71  & \textbf{0.75} & 0.53  & 0.43  & 0.34  & 0.00  & \textbf{0.79} & 0.60  & 0.50  & 0.39  & 0.74  & 0.00 \\
    \toprule
    \multicolumn{19}{c}{Black-Box} \\
    \midrule
      & \multicolumn{6}{c|}{Fashion-MNIST} & \multicolumn{6}{c|}{CIFAR-10} & \multicolumn{6}{c}{ImageNet} \\
    \midrule
      & Our & PD & DDN$_a$ & PGD$_a$ & TVM & $F$ & Our & PD & DDN$_a$ & PGD$_a$ & TVM & $F$ & Our & PD & DDN$_a$ & PGD$_a$ & TVM & $F$ \\
    \midrule
    Clean & 0.90  & 0.90  & 0.86  & 0.84  & 0.67  & \textbf{0.92} & 0.82  & 0.79  & 0.75  & 0.64  & 0.35  & \textbf{0.86} & 0.89  & 0.66  & 0.78  & 0.72  & 0.75  & \textbf{0.95} \\\hline
    FGSM-U  & \textbf{0.72} & 0.70  & 0.68  & 0.71  & 0.46  & 0.50  & \textbf{0.43} & 0.27  & 0.41  & 0.41  & 0.31  & 0.50  & \textbf{0.60} & 0.49  & 0.51  & 0.48  & 0.54  & 0.50 \\
    PGD-U & 0.78  & 0.80  & 0.77  & \textbf{0.82} & 0.48  & 0.50  & 0.66  & \textbf{0.70} & 0.68  & 0.58  & 0.31  & 0.50  & \textbf{0.63} & 0.61  & 0.58  & 0.50  & 0.51  & 0.50 \\
    PGD-T & 0.79  & 0.78  & 0.74  & \textbf{0.81} & 0.43  & 0.50  & 0.63  & \textbf{0.73} & 0.70  & 0.59  & 0.30  & 0.50  & \textbf{0.65} & 0.52  & 0.55  & 0.49  & 0.50  & 0.50 \\\hline
    Dfool-U & \textbf{0.87} & 0.86  & 0.84  & \textbf{0.87} & 0.48  & 0.50  & \textbf{0.78} & 0.76  & 0.71  & 0.61  & 0.29  & 0.50  & \textbf{0.67} & 0.60  & 0.58  & 0.51  & 0.43  & 0.50 \\
    CW-U  & \textbf{0.88} & 0.87  & 0.84  & 0.87  & 0.48  & 0.50  & \textbf{0.78} & 0.75  & 0.71  & 0.61  & 0.30  & 0.50  & \textbf{0.65} & 0.58  & 0.51  & 0.51  & 0.46  & 0.50 \\
    CW-T  & \textbf{0.87} & \textbf{0.87} & 0.84  & 0.85  & 0.53  & 0.50  & \textbf{0.77} & 0.75  & 0.71  & 0.60  & 0.29  & 0.50  & \textbf{0.67} & 0.45  & 0.56  & 0.51  & 0.44  & 0.50 \\
    DDN-U & \textbf{0.88} & 0.87  & 0.84  & 0.87  & 0.50  & 0.50  & \textbf{0.77} & 0.76  & 0.72  & 0.61  & 0.30  & 0.50  & \textbf{0.67} & 0.43  & 0.57  & 0.50  & 0.45  & 0.50 \\
    DDN-T & \textbf{0.88} & 0.87  & 0.84  & 0.87  & 0.49  & 0.50  & \textbf{0.77} & 0.74  & 0.71  & 0.60  & 0.28  & 0.50  & \textbf{0.68} & 0.36  & 0.53  & 0.46  & 0.41  & 0.50 \\
    \bottomrule
    \end{tabular}}%
  \label{tab:end2end}%
\end{table*}%


\subsection{End-to-end Performance of \name}

Here, we evaluate the end-to-end image classification performance of \name with the presence of four baselines. The first is PixelDefend, which can be used to detect and recover adversarial examples. The second is Total Variance Minimization (TVM), which is a type of adversarial example rectifying method~\cite{GuoRCM18}. The rest two are adversarial training methods based on PGD~\cite{madry2017pgd} and DDN~\cite{rony2019ddn}, denoted as $PGD_a$ and $DDN_a$, respectively. For the adversarial training baselines, we use adversarial examples to fine-tune the classifier, and then generate new adversarial examples to fool the fine-tuned classifier. Through iterating the training and attack process, the classifier is expected to converge to a robust state. Compared with our detector-rectifier framework, adversarial training is very time and computational resources consuming.

Table~\ref{tab:end2end} gives the comparative results of \name and the baselines. The performance is measured by the classification accuracy. The column $F$ corresponds to the classification accuracy of adversarial examples for an unprotected classifier. As can be seen,

$\bullet$ The accuracy is very low for the classifier $F$ under the grey-box attacks, and it even falls to zero under the $l_2$ attacks. In contrast, all the baselines can improve the classification accuracy and \name is the best.

$\bullet$ For clean examples, the accuracy of $F$ is better than others. The reason is that $F$ is directly trained on clean examples while our rectifier and the baselines are not. Nevertheless, the accuracy of \name is significantly higher than the baselines and is very close to that of $F$.

$\bullet$ In the black-box experiments, to ensure fair comparison under a same standard, we balance the proportion of successful and failed adversarial examples in the test set so as to fix the performance of $F$ to 50\%. The performance is calculated over the balanced test set. Again \name significantly outperforms $F$ and the baselines.

\begin{table}[t]\small
  \centering
  \caption{Classification accuracy of \name under white-box attacks.}
    \begin{tabular}{l|c|c|c}
    \toprule
    \multicolumn{4}{c}{X-Ensemble} \\
    \midrule
      & \multicolumn{1}{l|}{Fashion-MNIST} & \multicolumn{1}{l|}{CIFAR-10} & \multicolumn{1}{l}{ImageNet} \\
    \midrule
    PGD-T & 0.87 & 0.67 & 0.72 \\
    \midrule
    CW-T & 0.90 & 0.69 & 0.83 \\
    \midrule
    DDN-T & 0.90 & 0.71 & 0.78 \\
    \bottomrule
    \end{tabular}%
  \label{tab:white-box}%
   \vspace{-0.4cm}
\end{table}%
Finally, we test the performance of \name via white-box attacks. Since there exists no white-box algorithms that can directly attack \name, we propose a new attack approach based on Ref.~\cite{carlini2017adversarial}. Specifically, we concatenate all sub-detectors and the classifier $F$ as a new function and generate targeted adversarial examples to mislead the new function. The effect of the white-box attacker is to require all sub-detectors to give a benign prediction and require $F$ to give an incorrect prediction. If the white-box attack succeeds, an adversarial example can fool the RF ensemble by fooling all sub-detectors, and the classifier $F$ can also be misled at the same time. Table~\ref{tab:white-box} gives the classification accuracy of \name under the white-box attack. There are no untargeted attackers since the white-box approach can be only implemented using targeted algorithms. As the table shows, our model achieves very high classification accuracy by beating most of the white-box attacks. The reason is that it is very hard to search a perturbation that can fool all sub-detectors and the classifier $F$ simultaneously.


\section{Related Works}
\label{sec:relate}


\paratitle{Adversary Defense.} In the literature, there are two types of approaches to mitigate the threats of adversarial examples. The first is to directly strengthen classifiers against adversaries. The models along this line include three categories: $i)$ Adversarial Training~\cite{madry2017pgd,rony2019ddn}, which adds adversarial examples to augment training data and ensure a robust classifier. It usually consumes exorbitant time and computational resources by iteratively generating adversarial examples and optimizing classifiers, which makes it difficult to suit real-world large-scale applications. $ii)$ Modifying Model Architectures~\cite{BuckmanRRG18}, which manipulates model gradients to be nondifferentiable in inference. For instance, Decision-making Tree (DT) and Random Forest (RF) are ideal nondifferentiable models falling in this category, but they also suffer from relatively poor performance in image classification. In our \name model, we propose an RF-based ensemble framework to combine sub-detectors, which fully exploited the advantage of nondifferentiable models for adversarial examples detection. $iii)$ Rectifying Adversarial Examples, which removes input perturbations to some extent so that the input can be classified correctly, such as TVM~\cite{GuoRCM18}. However, traditional example rectifying methods do not consider the influences of pixels to attack effect when rectifying adversarial image pixels. Our \name model, in contrast, rectifies image pixels with the guidance of interpretation maps to ensure higher performance.

The second type of approaches for adversary defense focuses on the detection of adversarial examples. Models along this line mainly include using statistical methods to verify hand-crafted features of images and the network parameters~\cite{song2017pixeldefend,hu2019new,lee2018simple}, and using neural networks to judge whether an input is clean or adversarial~\cite{DBLP:conf/ccs/MengC17}. While neural network models are typically deemed more powerful in detection, they might suffer from adversarial attacks themselves. Our \name model solves this problem by adopting an RF-based ensemble framework. More importantly, \name plays a dual role as a detector-rectifier by establishing a novel joint framework for adversarial examples detection and rectification.

\paratitle{Neural Networks Interpretation.} There is a longstanding voice arguing that deep learning based systems are unexplainable black boxes and therefore cannot be used in crucial application domains such as medicine, investment and military. Many interpretation methods have been proposed to solve the black-box problem of DNNs, including time series analysis~\cite{wang2018multilevel},
intermediate representations~\cite{guidotti2019survey},
sensitivity analysis~\cite{sundararajan2017axiomatic}, and among others. As reported by Ref.~\cite{etmann2019connection}, the sensitivity analysis methods have close connection with adversarial robustness of neural networks.
Our \name model utilizes this information both in the detection part and rectification part to defend adversarial attacks.


\section{Conclusion}

In this paper, we proposed \name, a detection-rectification pipelined ensemble classifier for high-performance adversary defense. \name leverages target classifiers'  interpretation information of various types and designs interpreter-based multiple sub-detectors for accurate detection of adversarial examples. \name further employs the random forests model to form an ensemble detector upon sub-detectors, which not only improves the robustness of detectors against hybrid attacks but also avoids targeted attacks to detectors owing to the nondifferentiable property of a random forest classifier. We conducted extensive experiments to manifest the advantages of \name to competitive baseline methods under different attacks and rich scenarios.

\section*{ACKNOWLEDGMENTS}

This work was supported by the National Key R\&D Program of China (Grant No. 2019YFB2102100). Dr. Jingyuan Wang's work was partially supported by the National Natural Science Foundation of China (Grant No.61572059), the Fundamental Research Funds for the Central Universities (Grant No. YWF-20-BJ-J-839) and the CCF-DiDi Gaia Collaborative Research Funds for Young Scholars. Dr. Junjie Wu's work was partially supported by the National Special Program on Innovation Methodologies (Grant No. SQ2019IM4910001), and the National Natural Science Foundation of China (Grant No. 71531001, 71725002, U1636210, 71490723).

\bibliographystyle{ACM-Reference-Format}
\bibliography{ref}

\newpage
\newpage

\appendix

\section{Supplemental Materials}







\subsection{Datasets}
 We use the whole dataset of Fashion-MNIST~\cite{xiao2017fashion} and CIFAR-10~\cite{krizhevsky2009cifar}, but select images of 20 object classes from ImageNet~\cite{imagenet_cvpr09} as a subset for our experiments. All pixels are projected into [0, 1].





\vspace{-0.2cm}
\subsection{Model Structures}



Here we report the model structures in our experiments. For Fashion-MNIST and CIFAR-10, VGG11\cite{vgg11} structure is applied to the target classifier, detector and rectifier in \name, while VGG16\cite{vgg11} is for ImageNet experiments. The black-box classifiers for Fashion-MNIST , CIFAR-10 and ImageNet are contructed by CWnet\cite{carlini2017cw}, Wide-ResNet\cite{rony2019ddn} and ResNet152\cite{resnet} respectively.

Since it is very computational resource consuming to train a VGG16 and an ResNet152 model for ImageNet, we adopt the pre-trained VGG16 as the target classifier and the pre-trainded ResNet152 as the black-box classifier.
When using these models for ImageNet, we select the output logits of those 20 chosen classes out of the 1000 categaries and then compute the probabilities for them with $softmax$ function. In this way, the pre-trained VGG16 and ResNet152 both have more than 95\% accuracy on the ImageNet subset.

\vspace{-0.2cm}
\subsection{Setting of Attackers}

The codes of all attack methods in our experiments are implemented by AdverTorch \footnote{https://github.com/BorealisAI/advertorch} v0.2. 

The $\epsilon$ in FGSM and PGD is to constrain the $L_{\infty}$ perturbation of adversarial examples; the $\alpha$ in PGD and the learning rate ($lr$) in CW control the step size in each iteration when searching the adversarial perturbation; 
the iteration ($I$) limits how many times that an iterative attack can compute. $\epsilon=0.031$, $\alpha=0.0078$ and $lr=0.01$
for all the three datasets. $I$ is set to 100 for Fashion-MNIST and CIFAR-10, 50 for ImageNet.Other parameters use their default values in AdverTorch. 

OnePixel attack is only evaluated in unvaccinated training. We set the pixel numbers to 3, which means the attack can only modify 3 pixel values in an image. And its iteration is set to 30, with 100 candidates in each iteration.







\vspace{-0.2cm}
\subsection{Interpretation Methods}

VG, GBP and IG methods are implemented on our own and use LRP code from~\cite{montavon2019layer}. The integrated step in IG is set to 50. Note that LRP from~\cite{montavon2019layer} cannot be applied on ResNet152 directly. So we remove the LRP detector only for ImageNet.

\vspace{-0.2cm}
\subsection{Details of Rectifier}
We use Alg. \ref{algo:masked_image} to compute rectified images on adversarial examples and the hyperparameter $\alpha$ is set 0.6, 0.9, 0.5 for Fashion-MNIST, CIFAR-10 and ImageNet respectively.And we find that a rectifier tuned with rectified images of DDN-T mixed with clean images has  a better performance. 

\begin{algorithm} [t]\footnotesize
\caption{Rectified Image For Tuning Rectifier}\label{algo:masked_image}
\begin{algorithmic}[0]
  \STATE {\bf Variables:} $\{D_1,...,D_j\}$ are the sub-detectors that predict an input image $x$ as an adversarial one, $\{R_1,...,R_j\}$ are the interpreting methods corresponding to $\{D_1,...,D_j\}$ respectively, $\alpha \in (0,1)$ is a threshold parameter, $rand()$ returns a random value in $[0, 1]$, and $\sigma$ is the variance of pixel values in $x$.
  \FOR{$k=1$ to $j$ }
      \STATE $E_k \gets Entropy(D_k(x))$
  \ENDFOR
  \STATE $R \gets R_i$ where $i=argmin(E_1,...,E_j)$
  \STATE $g \gets R(x)$
  \STATE $thres \gets \alpha*(\max(g)-\min(g)) + \min(g)$
  \FOR{ ixel\ $(i,j)$ in $x$}
      \IF{$g_{i,j} > thres$ \textbf{and} $ rand() > 0.5$}
      \STATE $x_{i,j} \gets x_{i,j} + Normal(0, \sigma)$
      \ENDIF
  \ENDFOR
  \STATE \textbf{return} $x$
\end{algorithmic}
\end{algorithm}

\vspace{-0.2cm}
\subsection{Baselines}
In this section, we briefly introduce the sources of four baselines used in our work.

$\bullet$ \textbf{PD.} The code is from ~\cite{song2017pixeldefend}. When purifying images, $\epsilon$ is set to 0.125 for Fashion-MNIST and 0.063 for both CIFAR-10 and ImageNet. The pretrained model of PixelCNN for CIFAR-10 is from $openai^2$, and we train PixelCNN on Fashion-MNIST and ImageNet with code from $openai$ \footnote{https://github.com/openai/pixel-cnn}.

$\bullet$ \textbf{TWS.} The code is from \cite{hu2019new}. We set the parameters n\_radius = 0.01, targeted\_lr = 0.0005, t\_radius = 0.5, u\_radius = 0.5 and untargeted\_lr = 1.

$\bullet$ \textbf{MHL.} The code is from \cite{lee2018simple}. The magnitude of noise starts from 0.05 to 0.3 with an interval of 0.05 to compute its AUC.

$\bullet$ \textbf{TVM.} The code is from~\cite{tvm}. We set TVM\_WEIGHT = 0.03, PIXEL\_DROP\_RATE = 0.5, TVM\_METHOD = 'tvl2'.

$\bullet$ \textbf{Adversarial Training} Adversarial training is computational expensive. To speed up the training process while achieving a good accuracy, we use adversarial data to fine tune the original classifier with 15 epochs.

\vspace{-0.2cm}
\subsection{White-box Attacker for \name}
In this section, we show the details of the end-to-end white-box evaluation in our experiments.

Carlini \etal \cite{carlini2017adversarial} proposed a variant of CW attack to combine a classifier and a neural network detector into a new classifier $G$ with $L$+1 classes, whose ${(L+1)}^{th}$ class is for adversarial inputs. In this way, an attacker can directly attack $G$ to break $F$ and $D$ at the same time. $G$ is defined as,
\begin{equation}\small
    G(x)_i =
    \left\{ \begin{array}{lcr}
    F(x)_i &~~~ \mathrm{if}~~ i \leq L \\
    (D(x)+1) \cdot max_jF(x)_j &~~~ \mathrm{if}~~  i = L+1
     \end{array} \right.
     \label{equ:n+1classifier}
\end{equation}
where if $x$ is adversarial then $D(x) > 0$, so we have $G(x)_{L+1} > max(F(x))$ and  $argmax_i G(x)_i = L+1$; if $x$ is clean then $D(x) < 0$ , so we have $argmax_i G(x)_i = argmax_i F(x)_i$. An attacker will optimize this joint objective (target $t \neq l$ and $t \neq L+1$) to construct adversarial examples on $G$.

In our experiment, an attacker needs to fool all the detectors in \name. So we modify $G$ into an L+4 classifier as,
\begin{equation}\small
    G(x)_i =
    \left\{ \begin{array}{lcr}
    F(x)_i &~~~ \mathrm{if}~~ i \leq L \\
    (D_k(x) + 1) \cdot max_jF(x)_j &~~~ \mathrm{if}~~  i = L+k
     \end{array} \right.
     \label{equ:n+1classifier}
\end{equation}
where $D_k$ is one part of X-DET and $k=1,2,3,4$. Notice that here we remove the LRP detector since it is not differentiable in the second-order. So that a targeted attacker can generate examples on the classifier and the detectors to perform a white-box attack.

\end{document}